\documentclass[10pt,twocolumn,letterpaper]{article}

\usepackage[pagenumbers]{cvpr} 

\usepackage{graphicx}
\usepackage{amsmath}
\usepackage{amssymb}
\usepackage{booktabs}
\usepackage[accsupp]{axessibility} 

\usepackage[pagebackref,breaklinks,colorlinks]{hyperref}

\usepackage[capitalize]{cleveref}
\crefname{section}{Sec.}{Secs.}
\Crefname{section}{Section}{Sections}
\Crefname{table}{Table}{Tables}
\crefname{table}{Tab.}{Tabs.}

\def\cvprPaperID{29} 
\def\confName{CVPRW}
\def\confYear{2023}

\begin{document}

\title{Generating Adversarial Samples in Mini-Batches May Be Detrimental To Adversarial Robustness}

\author{Timothy Redgrave, Colton Crum\\
Department of Computer Science and Engineering University of Notre Dame \\
Notre Dame IN 46556, USA \\
{\tt\small tredgrav@nd.edu,ccrum@nd.edu}
}
\maketitle

\begin{abstract}
Neural networks have been proven to be both highly effective within computer vision, and highly vulnerable to adversarial attacks. Consequently, as the use of neural networks increases due to their unrivaled performance, so too does the threat posed by adversarial attacks. In this work, we build towards addressing the challenge of adversarial robustness by exploring the relationship between the mini-batch size used during adversarial sample generation and the strength of the adversarial samples produced. We demonstrate that an increase in mini-batch size results in a decrease in the efficacy of the samples produced, and we draw connections between these observations and the phenomenon of vanishing gradients. Next, we formulate loss functions such that adversarial sample strength is not degraded by mini-batch size. Our findings highlight a potential risk for underestimating the true (practical) strength of adversarial attacks, and a risk of overestimating a model's robustness. We share our codes to let others replicate our experiments and to facilitate further exploration of the connections between batch size and adversarial sample strength.

\end{abstract}

\section{Introduction}
Advances within neural networks have allowed them to achieve unprecedented levels of success across a number of computer vision tasks \cite{he2015delving,redmon2016you}. This in turn has led to their adoption across a wide range of applications, including within safety-critical systems such as facial recognition \cite{deng2019arcface} and autonomous vehicles \cite{kebria2019deep}. However, despite the impressive performance and remarkable potential displayed by neural networks, researchers have shown that these systems are highly susceptible to adversarial attacks - inputs which have been specifically modified in order to fool or mislead their targets into making incorrect decisions \cite{biggio2013evasion,szegedy2014intriguing}. Following this discovery, there has been a tremendous amount of research focused on developing methods which can be used to effectively train and assess adversarially robust models, leading to a number of developments on both fronts.

On the side of adversarial assessment, many advances have come from the development of newer and stronger attacks which represent a greater range of threat models. Early adversarial attacks focused on performing white-box attacks which leveraged gradient-based information in a relatively straightforward manner \cite{goodfellow2014explaining}, but more recent works have led to the development of stronger white-box attacks through iterative methods \cite{carlini2017towards,madry2018towards}, strong and efficient black-box attacks \cite{chen2020hopskipjumpattack,andriushchenko2020square
}, and even attacks which remain effective when deployed in a physical setting \cite{athalye2018synthesizing,komkov2021advhat}. Through the creation of additional and stronger attacks, it is now possible to test models against a larger amount of theoretical threat models, thereby allowing for a more thorough assessment of the vulnerabilities of these neural networks. 

Within the realm of adversarial robustness training, there have also been a number of significant developments. These include the development of principled formulations for adversarial training \cite{madry2018towards}, methods which allow for more efficient and easier implementations of adversarial training \cite{NEURIPS2019_7503cfac,wong2019fast}, and techniques which incorporate strategies such as smoothing in order to boost the efficacy of existing training methods \cite{chen2022efficient}.

However, despite the significant advances displayed in both adversarial training and testing, gaps still remain on both fronts. Often existing training techniques are evaluated against iterative white-box attacks which make use of the gradient, which can lead to an overestimation of model robustness through obfuscated gradients \cite{papernot2017practical,athalye2018obfuscated}. Likewise, it is difficult to assess the strength of novel adversarial attacks as this requires examining their efficacy against both baseline and adversarially trained models. These difficulties highlight the continued need to ensure not only the efficiency, but also the validity of claims made with regards to adversarial robustness. 

\begin{figure*}[t]
  \centering
  \includegraphics[width=0.975\linewidth]{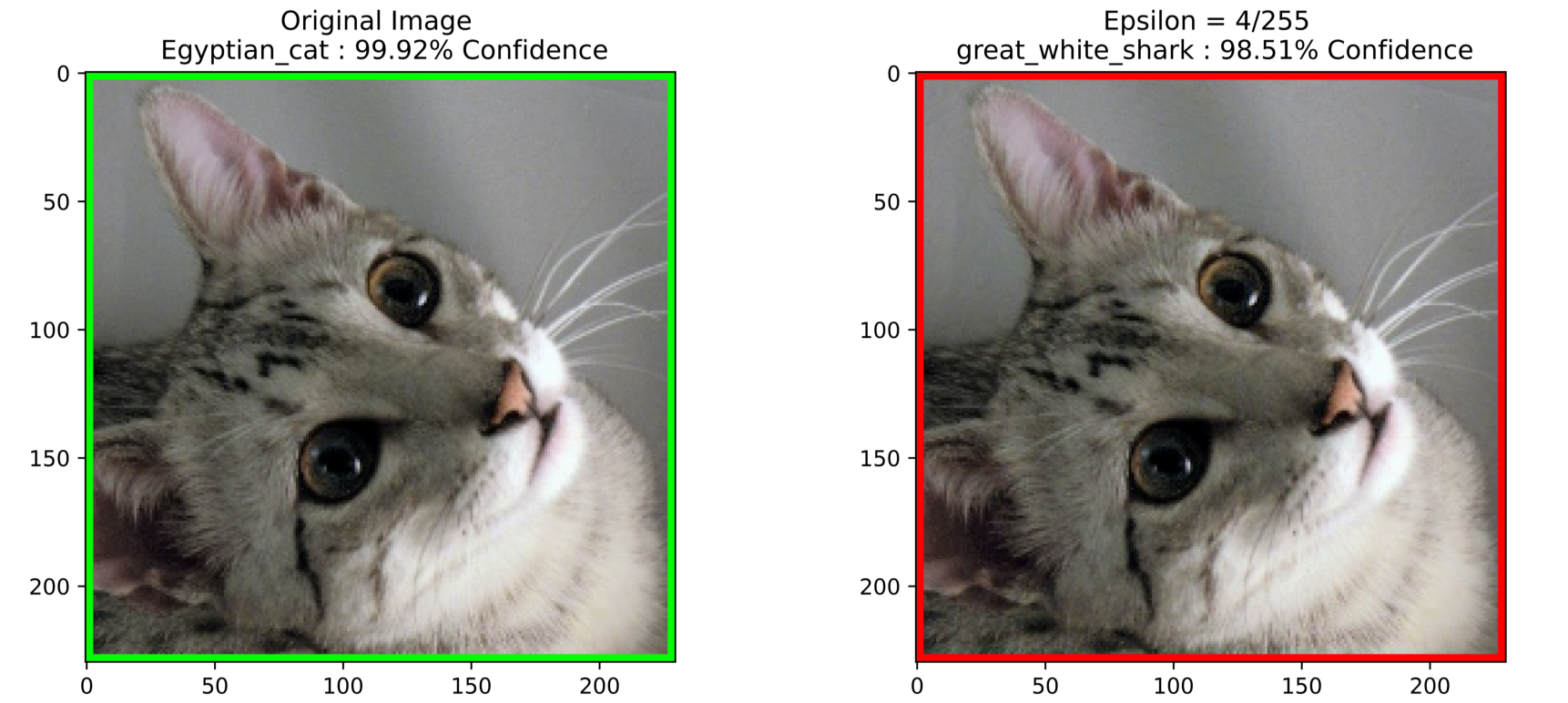}

   \caption{The image above shows the vulnerability of neural networks to adversarial attacks. Perturbing the input image by an \(\epsilon\) of just four pixel intensity levels (out of 255) within the \(L_{\infty}\) norm, an amount practically imperceptible to humans, is enough to cause the model to change the classification of the image from Egyptian cat to great white shark.}
   \label{fig:Adversarial_Example}
\end{figure*}

To this end, we identify a potential obstruction on the path to adversarial robustness - the generation of adversarial samples within large mini-batches. In this work we conduct experiments which demonstrate that the creation of adversarial samples within a batched setting leads to a degradation in the strength of the samples created. We attribute this effect to the reduction of the loss over a mini-batch to the mean loss, a process which results in vanishing gradients. We show that without this reduction, the negative effect of increased batch size is removed, and we discuss how this information can be leveraged moving forward. While our findings are relatively straightforward and align with an intuitive understanding of the issue, we are unaware of any previous works which explicitly address these concerns, and we argue that it is essential to highlight any potential mechanisms which may lead to a reduction in adversarial strength which could in turn lead to an underdevelopment or over-assessment of model robustness. The code used to generate the results for our project may be found here: \url{https://github.com/AdvCVPR23/Adversarial-Mini-Batching}.

\section{Preliminaries}
\subsection{Cross-Entropy Loss}
Cross-entropy loss is the most commonly used loss function when it comes to the supervised training of neural networks for classification tasks, and it plays a prominent role within a number of white-box adversarial attacks. For our purposes, we will consider the formulation for the cross-entropy loss for a mini-batch to be \begin{equation} \label{eq:Cross-Entropy}
    CE = -\frac{1}{N}\sum_{i=1}^{N} c_{n} \textrm{log} (\frac{e^{f(x_{n})_{c_{n}}}}{\sum_{c=1}^{C} e^{f(x_{n})_{c_{j}}}} )
\end{equation}
where \(N\) represents the number of samples in the mini-batch, \(C\) represents the number of classes within the given classification problem, \(c_{n}\) represents the class label for a given sample, and \(f(x_{n})\) represents the unnormalized logits of a given classifier \(f\) and a sample \(x_{n}\) (thereby making the \( \frac{e^{f(x_{n})_{c_{n}}}}{\sum_{c=1}^{C} e^{f(x_{n})_{c_{j}}}} \) term represent the softmax score for a given class \(c_{n}\) ). This formulation aligns with the default formulations given for cross-entropy loss within both TensorFlow and PyTorch, and for our purposes, it is most important to note the \(\frac{1}{N}\) term which gives us the mean loss rather than the total loss.

\subsection{Fast Gradient Method}
Introduced by Goodfellow \textit{et al.} \cite{goodfellow2014explaining}, the Fast Gradient Method (FGM) is one of the earliest and simplest methods that can be used to generate adversarial samples (within the original paper \cite{goodfellow2014explaining} the attack is applied only within the \( L_{\infty} \) norm and is called the Fast Gradient Sign Method (FGSM), however, when applied to other norms the sign operation is no longer used). FGM creates adversarial images in the following manner: first, a base image that is to be perturbed is fed through the model that is being attacked. The model's outputs are then used in conjunction with a given loss function in order to calculate the model's loss with respect to that particular image. This loss is then back-propagated to the input, wherein the resultant gradients are added to the image to create the adversarial image. In the final step of FGM, this adversarial image is clipped such that it is within some fixed distance $\epsilon$ of the original image (the distance $\epsilon$ will be dependent on what norm the target is operating within). Mathematically, this idea can be captured in the following equations: \begin{equation}
    x_{adv} = x + \epsilon \cdot \frac{\nabla_{x} L(f(x),y)}{||\nabla_{x} L(f(x),y)||_{2}}
\end{equation} for the \(L_{2}\) norm and \begin{equation}
    x_{adv} = x + \epsilon \cdot \textrm{Sign} ( \nabla_{x} L(f(x),y))
\end{equation} for the \(L_{\infty}\) norm respectively. Here \(x\) represents a clean or unperturbed sample, \(\nabla_{x} L(f(x),y) \) represents the gradient of the loss with to the input \(x\) and some given set of labels \(y\) (for our purposes, we consider cross-entropy loss as described in equation \ref{eq:Cross-Entropy}), and \(x_{adv}\) represents the newly constructed adversarial sample.

By modern standards, FGM is considered a simple or weak adversarial attack as it is a single-step method and only calculates the gradients with respect to the base image once and does not perform any randomization. However, despite the simplicity of this method, it remains highly effective against models that are trained in a non-robust manner, can be executed relatively quickly due to its single-step nature, and has seen surprising levels of success in some modern adversarial training regimes \cite{wong2019fast}. For these reasons we use FGM to represent a weak attack to test the effect of mini-batch size against.

\subsection{Projected Gradient Descent}
The second attack method we use within our assessment is Projected Gradient Descent (PGD) \cite{madry2018towards,tsipras2018robustness}. PGD operates by first injecting some level of noise into a given base image. PGD then iteratively performs the FGM attack using this modified base image in order to produce a stronger adversarial sample. This can be understood mathematically as \begin{equation}
    x_{adv}^{t+1} = \textrm{FGM}(x_{adv}^{t}), x_{adv}^{1} = \textrm{FGM}(x+noise)
\end{equation} with \(noise\) representing any type of noise bounded within the given \(\epsilon\) and \(x_{adv}^{t}\) representing the adversarial sample after \(t\) iterations of the FGM attack. The iterative nature of PGD combined with the random initialization allows for the construction of far stronger adversarial samples than single-step methods such as FGM, and PGD has been claimed to be the strongest first-order attack as a result \cite{madry2018towards}. Based on this, we use PGD to test the effect of mini-batch size on a strong adversary.

\section{Experimental Design}
\subsection{Experiments}
Within our work we perform four sets of experiments in order to elucidate the effect of mini-batch size within adversarial sample generation.
\begin{itemize}
    \item \textbf{Baseline:} Our first set of experiments involves generating samples using the loss formulation given within equation \ref{eq:Cross-Entropy}. The results from these runs demonstrate the effect of mini-batch size on adversarial strength when using the default formulations for loss accumulation over mini-batches (as reflected in the implementations for both TensorFlow and Pytorch).
    \item \textbf{Batch Correction:} The second set of experiments performed removes the term \(\frac{1}{N}\) from equation \ref{eq:Cross-Entropy}, resulting in: \begin{equation} \label{eq:Batch_Corrected_CE}
    CE = -\sum_{i=1}^{N} c_{n}\textrm{log}(\frac{e^{f(x_{n})_{c_{n}}}}{\sum_{c=1}^{C} e^{f(x_{n})_{c_{j}}}})\end{equation} These results allow us to directly attribute the cause of adversarial strength degradation to the commonly used reduction over mini-batch size (i.e., the use of the mean loss rather than total loss). It is worth noting that the computational complexities of equations \ref{eq:Cross-Entropy} and \ref{eq:Batch_Corrected_CE} are equivalent and thus this formulation does not incur any additional overhead in terms of memory usage or computation time.
    \item \textbf{Mixed Precision:} Our third set of experiments repeat our experiments using our original loss formulation given in equation  \ref{eq:Cross-Entropy} while using 16-bit floating point precision instead of 32-bit floating point precision for our tensors. This allows us to more concretely connect the degradation in adversarial attack strength to the phenomenon of vanishing gradients.
    \item \textbf{Batch Corrected Mixed Precision:} Our final set of experiments focuses on performing attacks using equation \ref{eq:Batch_Corrected_CE} and 16-bit precision floats. These experiments allow us to further confirm our understanding of the mechanisms behind our observations
\end{itemize}
All experiments are run eight times with mini-batches of 1, 2, 4, 8, 16, 32, 64, and 128 images.

\subsection{Attack Parameters}
For each experiment we perform FGM and PGD within the Euclidean (\(L_{2}\)) and infinite (\(L_{\infty}\)) norms using the parameters given in table \ref{table:attacks_table}.

\begin{table}[h]
  \centering
    \begin{tabular}{ |c|c|c|c| }
    \hline
    Attack Type & Epsilon & Steps & Step Size\\
    \hline
    FGM \(L_{2}\) & 128/256 & 1 & \(\epsilon\) \\
    FGM \(L_{\infty}\) & 8/256 & 1 & \(\epsilon\) \\
    PGD \(L_{2}\) & 128/256 & 32 & \(2\cdot\epsilon/32\) \\
    PGD \(L_{\infty}\) & 8/256 & 32 & \(2\cdot\epsilon/32\) \\
    \hline
    \end{tabular}
    \caption{Attacks with a strength of \(\epsilon=8/256\) correspond to a strength of \(0.031\) in the normalized \([0,1]\) space as used in works such as \cite{athalye2018obfuscated}.}
    \label{table:attacks_table}
\end{table}
Performing both FGM and PGD allows us to assess the effect of mini-batch size on weak attacks and strong attacks respectively, and performing attacks across both the \(L_{2}\) and \(L_{\infty}\) norms allows us to assess the effect of mini-batch size across different measures of visual distance.

\begin{figure*}[ht]
    \subfloat[InceptionV3]{
        \includegraphics[width=0.4925\linewidth]{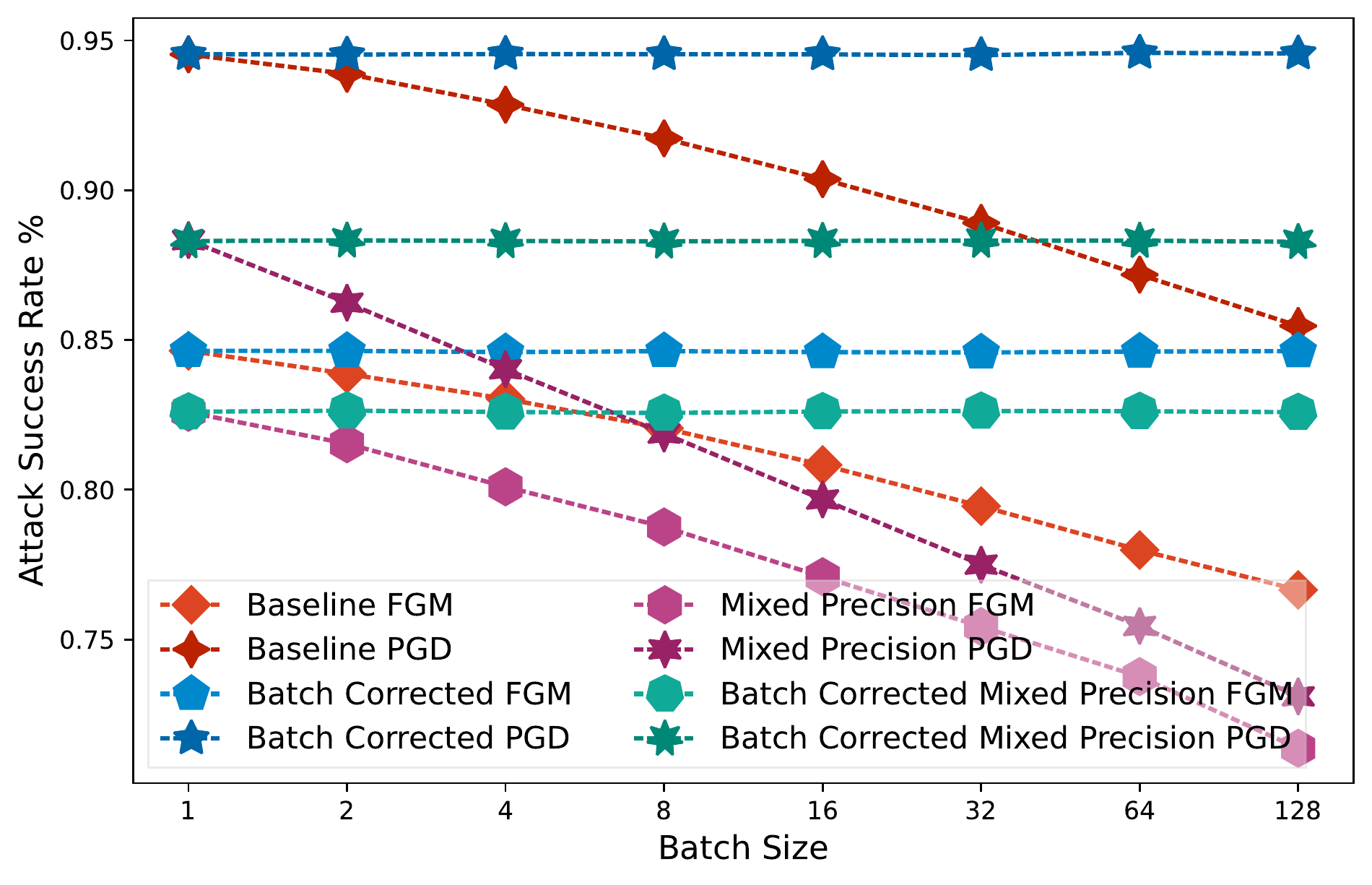}%
        \label{subfig:Inception_L2}%
    }\hfill
  \subfloat[Xception]{
      \includegraphics[width=0.4925\linewidth]{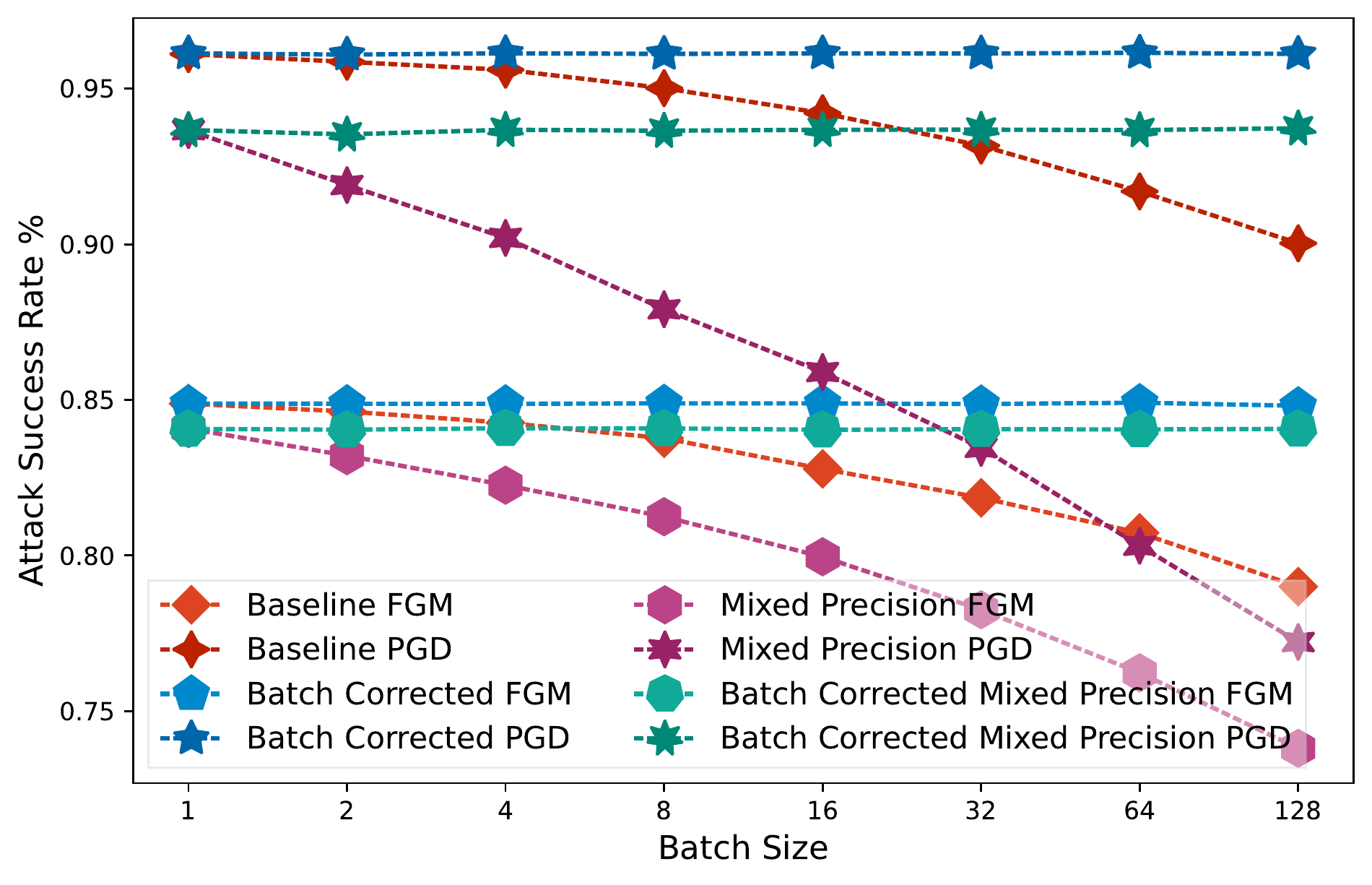}%
      \label{subfig:Xception_L2}%
    }\hfill\\
    \subfloat[ResNet50]{
        \includegraphics[width=0.4925\linewidth]{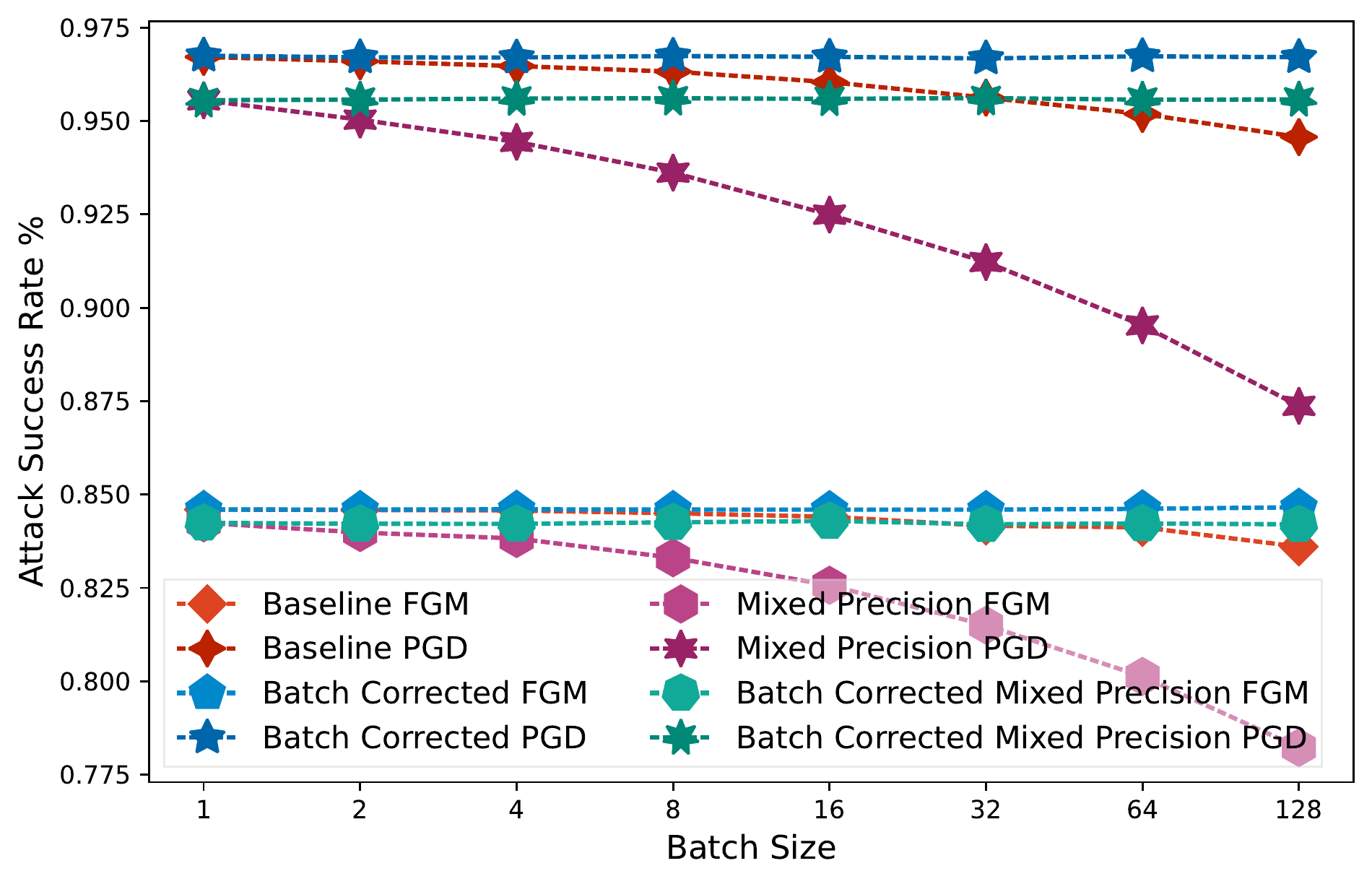}%
        \label{subfig:ResNet_L2}%
    }\hfill
  \subfloat[DenseNet121]{
      \includegraphics[width=0.4925\linewidth]{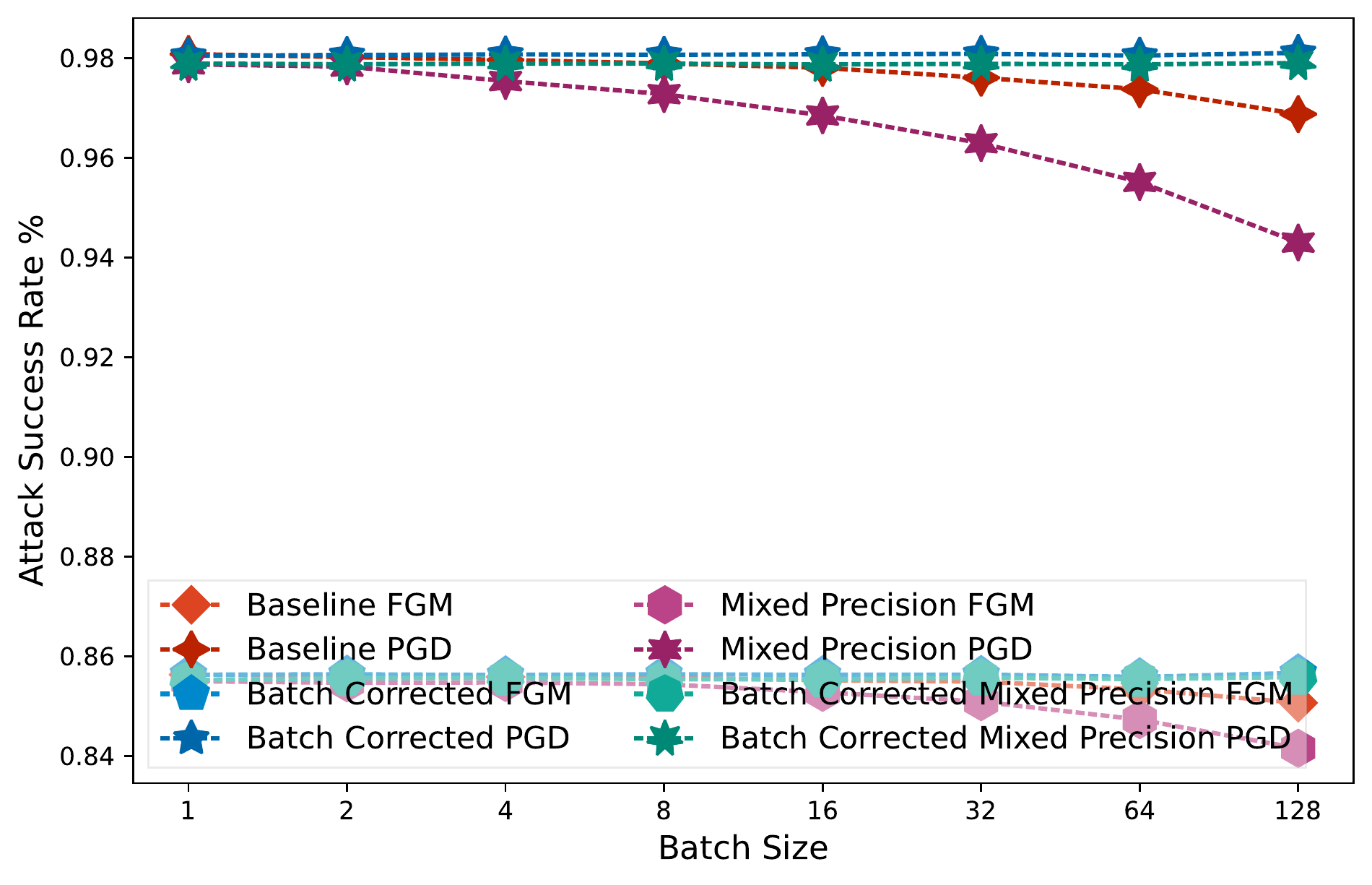}%
      \label{subfig:DenseNet_L2}%
  }
  \centering\caption {The graphs above show the detrimental effect of large mini-batch sizes when generating adversarial attacks bounded within the Euclidean norm (\(L_{2}\)). Lines with convex polygons represent FGM attacks, and lines with concave polygons show PGD attacks.}
  \label{fig:L2-Results}
\end{figure*}
\subsection{Dataset and Models}
In our experiments we make use of ImageNetV2 \cite{recht2019imagenet}. ImageNetV2 is a collection of 10,000 images consisting of 10 sample images for each of the 1000 different classes found within the original ImageNet \cite{deng2009imagenet}. While ImageNetV2 was primarily designed as an additional test set for models trained on the original ImageNet, ImageNetV2 \cite{recht2019imagenet} forms an ideal basis for testing adversarial image generation as it properly mirrors a real-world threat model wherein attackers may be able to manipulate genuine data which was not available to the model designers during the training or testing of a given model. For our experiments all images are resized to be 224x224.

We ran all experimental setups using four separate model architectures: InceptionV3 \cite{szegedy2016rethinking}, Xception \cite{chollet2017xception}, ResNet50 \cite{he2016deep}, and DenseNet121 \cite{huang2017densely}. Each model was pre-trained on ImageNet \cite{deng2009imagenet} and made use of the default TensorFlow weights and preprocessing functions (specific to each model).

\subsection{Additional Considerations}
In order to limit potential confounding factors within our analysis, we use an Intexl Xeon Gold 6226 CPU paired with an NVIDIA Quadro RTX 6000 GPU for all of our experiments. Additionally, in order to minimize the effect of the random initialization within PGD attacks, we repeat all runs five times and take the average result across the runs.

\section{Evaluation and Analysis}
\begin{figure*}[ht]
    \subfloat[InceptionV3]{
        \includegraphics[width=0.4925\linewidth]{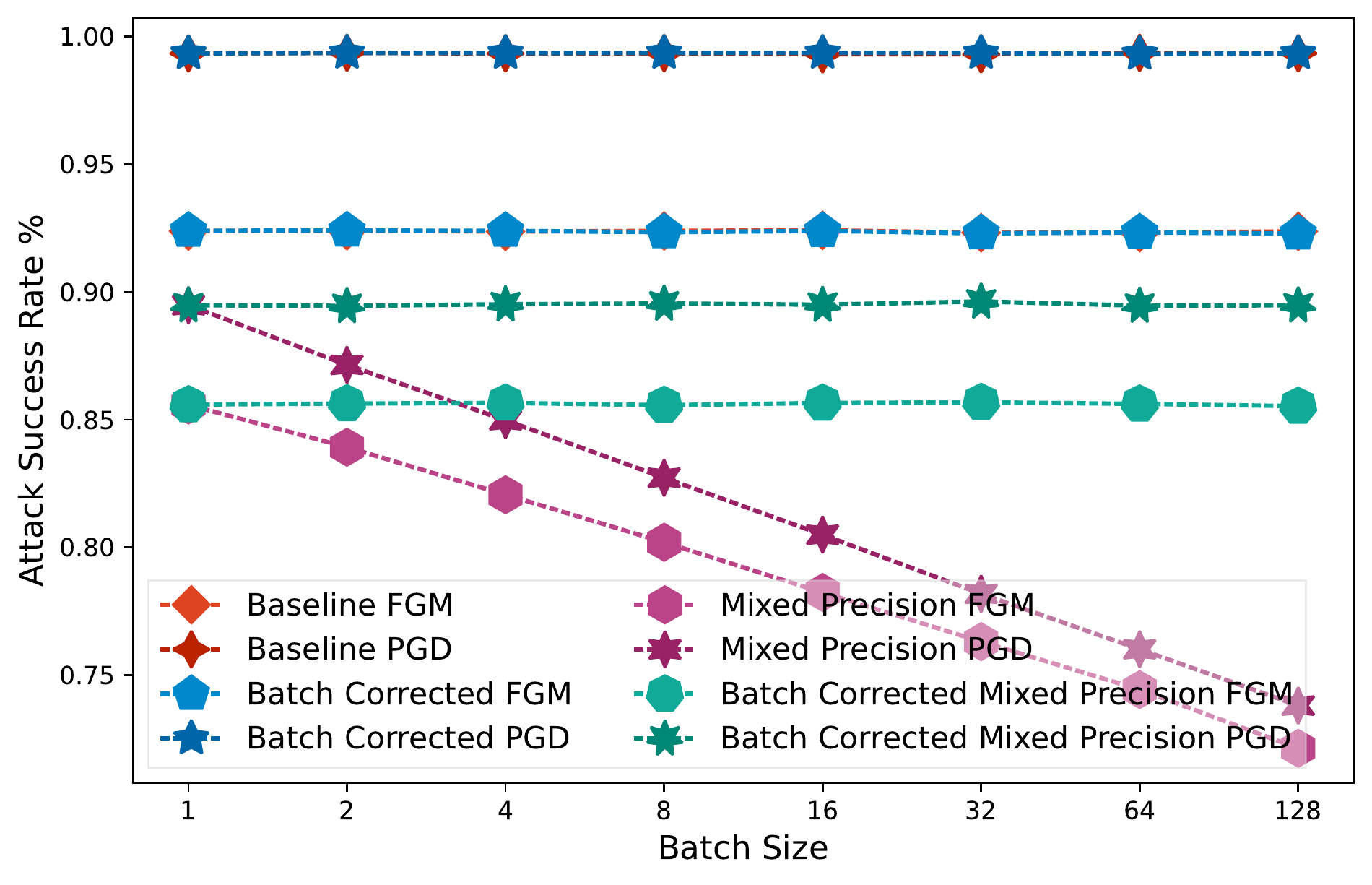}%
        \label{subfig:Inception_L-INF}%
    }\hfill
  \subfloat[Xception]{
      \includegraphics[width=0.4925\linewidth]{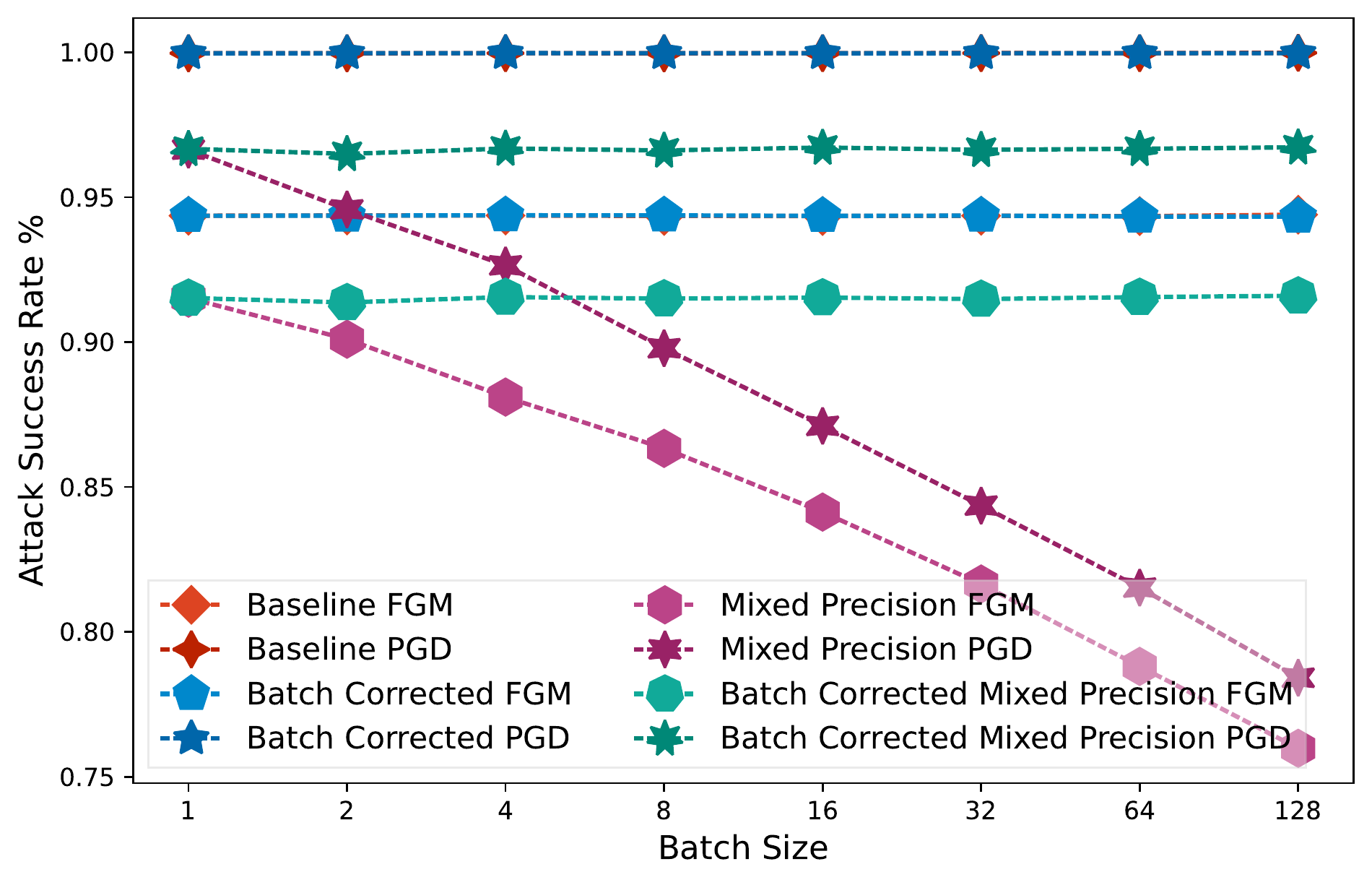}%
      \label{subfig:Xception_L-INF}%
    }\hfill\\
    \subfloat[ResNet50]{
        \includegraphics[width=0.4925\linewidth]{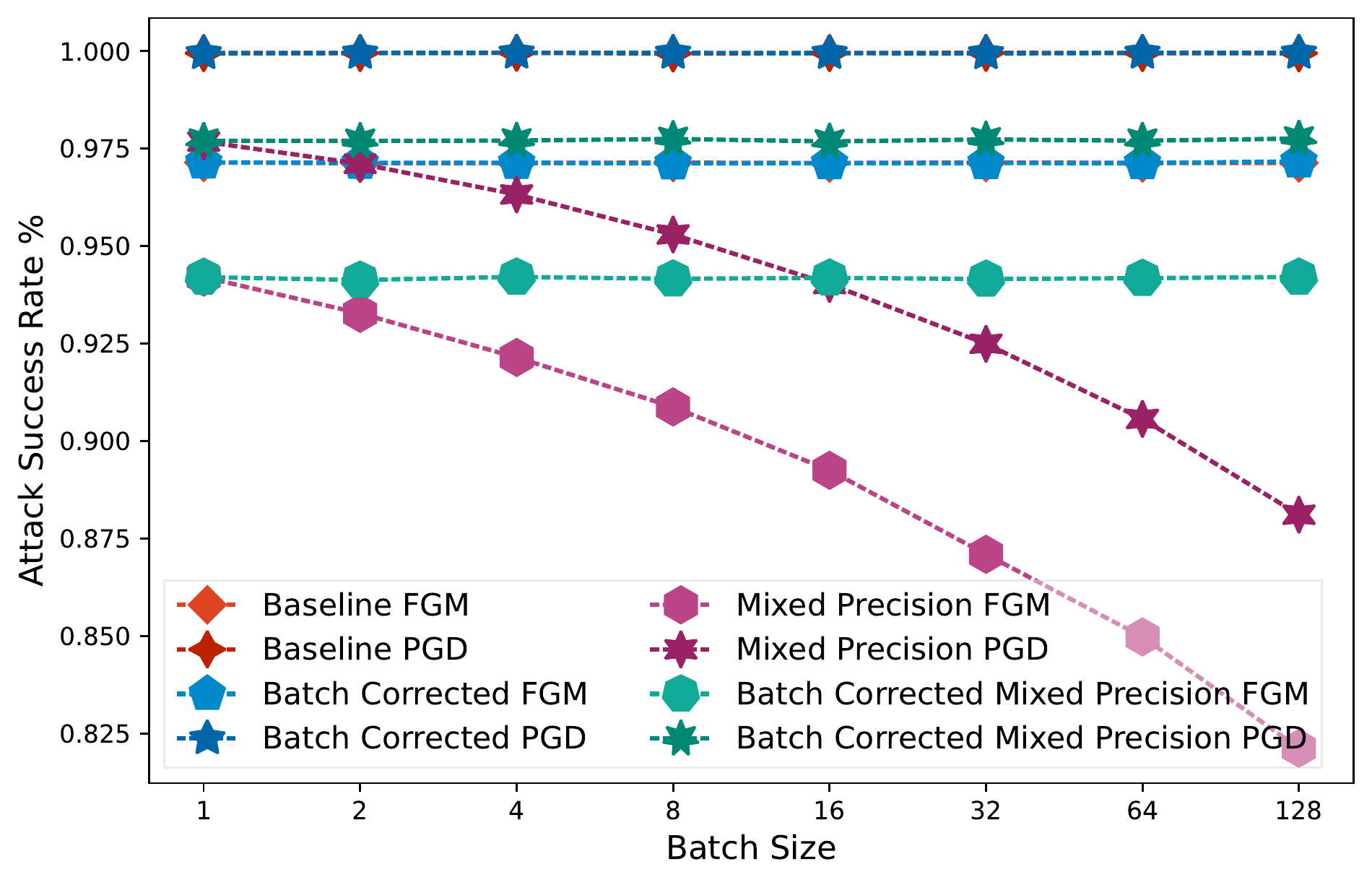}%
        \label{subfig:ResNet_L-INF}%
    }\hfill
  \subfloat[DenseNet121]{
      \includegraphics[width=0.4925\linewidth]{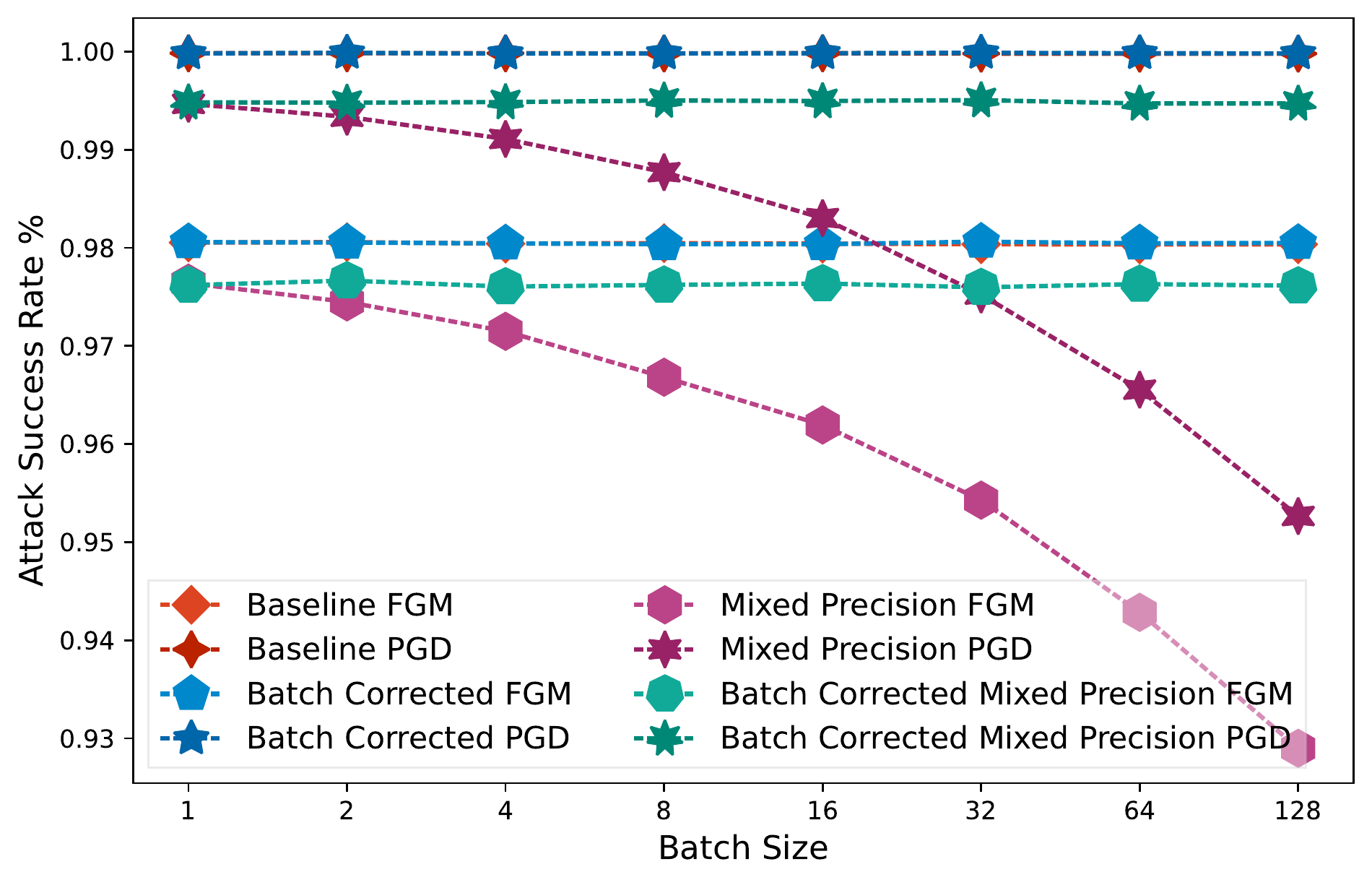}%
      \label{subfig:DenseNet_L-INF}%
  }
  \centering\caption {This figure shows the same set of results as fig. \ref{fig:L2-Results} but for attacks bounded within the infinite norm (\(L_{\infty}\)).
  }
  \label{fig:L-INF-Results}
\end{figure*}

\subsection{Baseline vs. Batch Correction}
The first thing which may be observed when examining the results of our experiments is the universal degradation of adversarial sample strength across attacks and models when using large mini-batches in conjunction with the proposed baseline cross-entropy loss formulation given in equation \ref{eq:Cross-Entropy}. While this effect is much more pronounced within attacks bounded within the Euclidean norm and for the InceptionV3 and Xception architectures (in contrast with the infinite norm and the ResNet50 and DenseNet121 architectures respectively), the use of larger mini-batch sizes during the generation of adversarial samples consistently leads to a decrease in the number of samples able to successfully fool the given classifier. The most drastic example of this can be seen when we look at the InceptionV3 model. When generating adversarial samples via PGD bound within the Euclidean norm (\(L_{2}\)), the rate of successful attacks against InceptionV3 drops from 94.54\% when samples are generated individually, to 85.47\% when samples are generated within batches of size 128.

In contrast, when we use our formulation given within equation \ref{eq:Batch_Corrected_CE} which calculates the total loss rather than the mean loss, we can see that the penalty to adversarial strength induced by larger batch sizes is alleviated. Under this corrected formulation attacks are able to retain their strength whether they are generating individual samples or they are generating samples in large batches. These findings align well with an intuitive view of the problem; if samples are generated based on the mean loss rather than the total loss, the contribution of each individual sample will decrease as the number of samples considered increases. This in turn will lead to gradients which are smaller in magnitude, and thus will incidentally result in vanishing gradients as larger and larger batch sizes are explored.

\subsection{Confirmation through Mixed Precision}
These results regarding the relationship between batch size and adversarial sample strength are further supported when we examine our mixed precision results. As can be observed from our findings, regardless of batch size, a decrease in precision leads to a drop in adversarial strength. This is to be expected as by halving our precision, we have a less granular view of our loss surface, and thus the descent direction given by the gradient will be less exact. The more interesting observation, however, arises when we examine our baseline mixed precision results and compare them against our batch corrected results. If we do this, we can  obverse that while adversarial samples generated in a mixed precision manner may result in a lower degree of adversarial strength for small batch sizes, as batch size increases, batch corrected samples generated in a mixed precision manner often reach a point where they are more effective on average than samples generated with full precision but without batch correction. This highlights the importance of using the sum of the loss over the mini-batch rather than the mean of the loss over the mini-batch when generating adversarial samples.

\subsection{Discussion}
Given the relative simplicity of our results and how well they align with what may be expected from an intuitive examination of the problem, it is worth asking whether these results are important. To that end, we present a two-fold argument as to the relevance of this work. The first argument comes from a desire to standardize robustness testing and verification. While a number of previous works (such as \cite{athalye2018obfuscated,carlini2017towards}) have discussed the need to build towards a shared set of baseline testing standards, to our knowledge our work is the first which explicitly discusses how to formulate loss functions such that sample strength is not degraded by mini-batch size during testing. This allows us to take greater advantage of parallelized environments without sacrificing any potential adversarial strength (and as a consequence, over estimating adversarial robustness). 

Our second argument stems from a slightly less obvious use case - the generation of adversarial samples during adversarial training. While switching from using the mean loss over a mini-batch to using the total loss over a mini-batch may be a relatively painless adjustment to make during model assessment, this switch is not as straightforward during model training. Mean loss reduction allows for a more direct comparison of models trained using different batch sizes, a smoother approximation of the loss surface, and generally more convenient numerical properties. In contrast, if the reduction of adversarial samples is done by the sum of the losses over a mini-batch, a number of hyper-parameters may need to be adjusted based on the batch size used including step-size, momentum decay, and more. Therefore, we can observe that even while the challenge of correcting for the effect of large batch sizes may be relatively easily addressed within model testing, it is a much more complicated issue to address during model training. Furthermore, while an incorrect formulation of the adversarial loss may lead to an over estimation of robustness if done during model testing, this same mistake may lead to a reduction in true model robustness if made during model training. As such, we assert that it is crucial to explicitly establish the potential consequences of incorrect adversarial training and assessment, and to build our training and evaluation tools such in such a way that these risks are mitigated. 

\section{Conclusion}
In this work we demonstrated the link between large mini-batch sizes and the reduction of adversarial strength within the adversarial samples created. Through testing formulations which used mean loss, total loss, and mixed precision, we were able to isolate the effect of batching within our analysis and demonstrate that while its magnitude may change based on the model or attack used, its negative impact remains present regardless. We then briefly discussed how this may affect models during both training and assessment, and we further affirm the need for standard testing tool and parameters for evaluating model robustness.

{\small
\bibliographystyle{ieee_fullname}
\bibliography{main}
}

\end{document}